\documentclass[11pt,a4paper]{article}
\usepackage[hyperref]{eacl2021}
\usepackage{tcolorbox}
\usepackage{times}
\usepackage{float}
\usepackage{latexsym}
\usepackage{amsmath}
\usepackage{graphics}
\usepackage{lipsum}
\usepackage{xurl}
\usepackage{color}
\usepackage{soul}
\usepackage{booktabs,chemformula}

\DeclareMathOperator*{\argmax}{argmax} % thin space, limits underneath in displays

\usepackage{microtype}

\aclfinalcopy % Uncomment this line for the final submission

\title{Extending Neural Keyword Extraction with TF-IDF tagset matching}
\author{Boshko Koloski \\
  Jo\v{z}ef Stefan Institute \\
  Jo\v{z}ef Stefan IPS\\
  Jamova 39, Ljubljana\\
  \texttt{boshko.koloski@ijs.si} \\\And
  Senja Pollak \\
  Jo\v{z}ef Stefan Institute\\Jamova 39, Ljubljana \\  \texttt{senja.pollak@ijs.si} \\\AND
   Bla\v{z}  \v{S}krlj \\
  Jo\v{z}ef Stefan Institute \\
    Jo\v{z}ef Stefan IPS\\Jamova 39, Ljubljana\\
    \texttt{blaz.skrlj@ijs.si}\\\And
  Matej Martinc \\
  Jo\v{z}ef Stefan Institute\\ Jamova 39, Ljubljana\\  \texttt{matej.martinc@ijs.si} \\
  }

\date{}

\begin{document}
\maketitle

\begin{tcolorbox}

 The final formatted version of this publication was published in  Proceedings of the EACL Hackashop on News Media Content Analysis and Automated Report Generation (EACL 2021), Online, April, 2021 and is available online at \url{https://www.aclweb.org/anthology/2021.hackashop-1.4}.

\end{tcolorbox}
\begin{abstract}
      Keyword extraction is the task of identifying words (or multi-word expressions) that best describe a given document and serve in news portals to link articles of similar topics. In this work, we develop and evaluate our methods on four novel data sets covering less-represented, morphologically-rich languages in European news media industry (Croatian, Estonian, Latvian, and Russian). First, we perform evaluation of two supervised neural transformer-based methods, Transformer-based Neural Tagger for Keyword Identification (TNT-KID) and Bidirectional Encoder Representations from Transformers (BERT) with an additional Bidirectional Long Short-Term Memory Conditional Random Fields (BiLSTM CRF) classification head, and compare them to a baseline Term Frequency - Inverse Document Frequency (TF-IDF) based unsupervised approach. Next, we show that by combining the keywords retrieved by both neural transformer-based methods and extending the final set of keywords with an unsupervised TF-IDF based technique, we can drastically improve the recall of the system, making it appropriate for usage as a recommendation system in the media house environment. 
\end{abstract}
\section{Introduction}
\label{sec:intro}

Keywords are words (or multi-word expressions) that best describe the subject of a document, effectively summarise it and can also be used in several document categorization tasks. In online news portals, keywords help with efficient retrieval of articles when needed. Similar keywords characterise articles of similar topics, which can help editors to link related articles, journalists to find similar articles and readers to retrieve articles of interest when browsing the portals. For journalists manually assigning tags (keywords) to articles represents a demanding task, and high-quality automated keyword extraction shows to be one of components in news digitalization process that many media houses seek for.

The task of keyword extraction can generally be tackled in an unsupervised way, i.e., by relying on frequency based statistical measures \citep{campos2018yake} or graph statistics \citep{vskrlj2019rakun}, or with a supervised keyword extraction tool, which requires a training set of sufficient size and from appropriate domain. While supervised methods tend to work better due to their ability to adapt to a specifics of the syntax, semantics, content, genre and keyword assignment regime of a specific text \citep{martinc2020tnt}, their training for some less resource languages is problematic due to scarcity of large manually annotated resources. For this reason, studies about supervised keyword extraction conducted on less resourced languages are still very rare. To overcome this research gap, in this paper we focus on supervised keyword extraction on three less resourced languages, Croatian, Latvian, and Estonian, and one fairly well resourced language (Russian) and conduct experiments on data sets of media partners in the EMBEDDIA project\footnote{http://embeddia.eu/}. The code for the experiments is made available on GitHub under the MIT license\footnote{\url{https://github.com/bkolosk1/Extending-Neural-Keyword-Extraction-with-TF-IDF-tagset-matching/}}.

In media house environments, automatic keyword extraction systems are expected to return a diverse list of keyword candidates (of constant length), which is then inspected by a journalist who manually selects appropriate candidates. While the state-of-the-art supervised approaches in most cases offer good enough precision for this type of usage as a recommendation system, the recall of these systems is nevertheless problematic. Supervised systems learn how many keywords should be returned for each news article on the gold standard train set, which generally contains only a small amount of manually approved candidates for each news article. For example, among the datasets used in our experiments (see Section \ref{sec:dataset}), the Russian train set contains the most (on average 4.44) present keywords (i.e., keywords which appear in the text of the article and can be used for training of the supervised models) per article, while the Croatian test set contains only 1.19 keywords per article. This means that for Croatian, the model will learn to return around 1.19 keywords for each article, which is not enough.

To solve this problem we show that we can improve the recall of the existing supervised keyword extraction system by:

\begin{itemize}
    \item Proposing an additional TF-IDF tagset matching technique, which finds additional keyword candidates by ranking the words in the news article that have appeared in the predefined keyword set containing words from the gold standard train set. The new hybrid system first checks how many keywords were returned by the supervised approach and if the number is smaller than needed, the list is expanded by the best ranked keywords returned by the TF-IDF based extraction system.
    \item Combining the outputs of several state-of-the-art supervised keyword extraction approaches. 
\end{itemize}

The rest of this work is structured as follows: Section \ref{sec:related-work} presents the related work, while Section \ref{sec:dataset} describes the datasets on which we evaluate our method. Section \ref{sec:methodology} describes our proposed method with all corresponding steps. The experiment settings are described in Section \ref{sec:media-experiments} and the evaluation of the proposed methods is shown in Section \ref{sec:eval}. The conclusions and the proposed further work are presented in Section \ref{sec:conc}.  
    
\section{Related Work}
\label{sec:related-work}

Many different approaches have been developed to tackle the problem of extracting keywords. The early approaches, such as KP-MINER \citep{el2009kp} and RAKE \citep{rose2010automatic} rely on unsupervised techniques which employ frequency based metrics for extraction of keywords from text. Formally, aforementioned approaches search for the words $w$ from vocabulary $\mathcal{V}$ that maximize a given metric $h$ for a given text $t$: 
\begin{equation*}
    \textrm{kw} = \argmax_{w \in \mathcal{V}} h(w,t). 
\end{equation*}
In these approaches, frequency is of high relevance and it is assumed that the more frequent a given word, the more important the meaning this word carries for a given document. Most popular such metrics are the na\"ive frequency (word count) and the term frequency-inverse document frequency (TF-IDF) \cite{tfidf}. 

Most recent state-of-the-art statistical approaches, such as YAKE \citep{campos2018yake}, also employ frequency based features, but combine them with other features such as casing, position, relatedness to context and dispersion of a specific term in order to derive a final score for each keyword candidate.

Another line of research models this problem by exploiting concepts from graph theory. Approaches, such as TextRank \citep{mihalcea2004textrank}, Single Rank \citep{wan2008single}, TopicRank \citep{bougouin2013topicrank} and Topical PageRank \citep{sterckx2015topical} build a graph $G$, i.e., a mathematical construct described by a set of vertexes $V$ and a set of edges $E$ connecting two vertices. In one of the most recent approaches called RaKUn \cite{vskrlj2019rakun}, a directed graph is constructed from text, where vertexes $V$ and two words $w_{i}, w_{i+1}$ are linked if they appear following one another. Keywords are ranked by a shortest path-based metric from graph theory - the load centrality.
   
The task of keyword extraction can also be tackled in a supervised way. One of the first supervised approaches was an algorithm named KEA \citep{witten2005kea}, which uses only TF-IDF and the term's position in the text as features for term identification. More recent neural approaches to keyword detection consider the problem as a sequence-to-sequence generation task \cite{meng2017deep} and employ a generative model for keyword prediction with a recurrent encoder-decoder framework and an attention mechanism capable of detecting keywords in the input text sequence whilst also potentially finding keywords that do not appear in the text. 

Finally, the newest branch of models consider keyword extraction as a sequence labelling task and tackle keyword detection with transformers. \citet{sahrawat2019keyphrase} fed contextual embeddings generated by several transformer models (BERT \citep{devlin2018bert}, RoBERTa \citep{liu2019roberta}, GPT-2 \citep{radford2019gpt2}, etc.) into two types of neural architectures, a bidirectional Long short-term memory network (BiLSTM) and a BiLSTM network with an additional Conditional random fields layer (BiLSTM-CRF). \citet{sun2020joint} on the other hand proposed BERT-JointKPE that employs a chunking network to identify phrases and a ranking network to learn their salience in the document. By training BERT jointly on the chunking and ranking tasks the model manages to establish balance between the estimation of keyphrase quality and salience.

Another state-of-the-art transformer based approach is TNT-KID (Transformer-based Neural Tagger for Keyword Identification) \cite{martinc2020tnt}, which does not rely on pretrained language models such as BERT, but rather allows the user to train their own language model on the appropriate domain. The study shows that smaller unlabelled domain specific corpora can be successfully used for unsupervised pretraining, which makes the proposed approach easily transferable to low-resource languages. It also proposes several modifications to the transformer architecture in order to adapt it for a keyword extraction task and improve performance of the model. 

\section{Data Description}
\label{sec:dataset}
%The experiments we conducted based on three different language branches. Estonian represented the Uralic language branch, Latvian represented the Baltic language branch, while the Slavic language branch was represented by the Croatian and the Russian language. The Estonian and Russian datasets contained news from the Ekspress Group from the Ekspress Media (ExM). The Latvian dataset was 

%\hl{Matej/Boshko - prosim preveri, kako je bil Ruski dataset zgrajen - ruske novice iz obeh datasetov ali le iz Estonskega}\textcolor{green}{Samo iz estonskega mensezdi}\hl{BK: Samo Estonski CONFIRMED}. The Croatian dataset was acquired from 24sata news portal belonging to Styria Media Group, one of the leading media groups in Austria, Croatia, and Slovenia. 
%OLD-MOVED: 

We conducted experiments on datasets containing news in four languages; Latvian, Estonian, Russian, and Croatian. Latvian, Estonian and Russian datasets contain news from the Ekspress Group, specifically from Estonian Ekspress Meedia (news in Estonian and Russian) and from Latvian Delfi (news in Latvian and Russian). The Croatian dataset was acquired from 24sata news portal belonging to Styria Media Group, one of the leading media groups in Austria, Croatia, and Slovenia. The dataset statistics are presented in Table \ref{tab:media-dataset}, and the datasets \cite{HackashopResources2021} and their train/test splits\footnote{\url{https://www.clarin.si/repository/xmlui/handle/11356/1403}} are publicly available. 
The media-houses provided news articles from 2015 up to the 2019. We divided them into training and test sets. For the Latvian, Estonian, and Russian training sets, we used the articles from 2018, while for the test set the articles from 2019 were used. For Croatian, the articles from 2019 are arranged by date and split into training and test (i.e., about 10\% of the 2019 articles with the most recent date) set. In our study, we also use tagsets of keywords. Tagset corresponds either to a collection of keywords maintained by editors of a media house (see e.g. Estonian tagset), or to a tagset constructed from assigned keywords from articles available in the training set. The type of tagset and the number of unique tags for each language are listed in Table \ref{tab:media-tagset}.

\begin{table}[h]
    \centering
        \begin{tabular}{| c | c | c |}
        \hline
        \textbf{Dataset} & \textbf{Unique tags} & \textbf{Type of tags}\\ \hline
        Croatian & 21,165 & Constructed  \\ 
        Estonian & 52,068 & Provided \\ 
        Russian & 5,899 &  Provided  \\  
        Latvian & 4,015 & Constructed  \\ \hline

    \end{tabular}
    \caption{Distribution of tags provided per language. The media houses provided tagsets for Estonian and Russian, while the tags for Latvian and Croatian were extracted from the train set.}
    \label{tab:media-tagset}
\end{table}

\begin{table*}[h]
    \centering
   %\hl{a lahko v 1. vrstici razbijes na 2 vrstici imena stolpcev, npr. Avg. train (zgoraj) in doc. len. (spodaj), s tem bo postala vecja berljivost stevilk}}
    \resizebox{\linewidth}{!}{
        \begin{tabular}{| c | c | c | c | c | c | c | c | c | c | c | c | c |}
        \cline{4-13}
         \multicolumn{3}{c}{}& \multicolumn{5}{|c|}{ \textbf{Avg. Train}} & \multicolumn{5}{|c|}{\textbf{Avg. Test}} \\ \hline

         \textbf{Dataset} & \textbf{Total docs} &  \textbf{Total kw.} & \textbf{Total docs}   & \textbf{Doc len} &  \textbf{Kw.} & \textbf{\% present kw.} & \textbf{present kw.} & \textbf{Total docs} & \textbf{Doc len} & \textbf{Kw.} & \textbf{\% present kw.} & \textbf{Present kw.}   \\ \hline
       % Croatian & 52,756 & 26,896 & 47,479 & 420.32 & 3.10 & 0.47 & 1.32 & 5,277 &  464.14 & 3.28 & 0.55 & 1.62   \\ 
        Croatian & 35,805 & 126,684 & 32,223 & 438.50 & 3.54 & 0.32 & 1.19 & 3582 & 464.39 & 3.53 & 0.34 & 1.26   \\ 

        Estonian & 18,497 & 59,242 & 10,750 & 395.24 & 3.81 &  0.65 & 2.77 & 7,747 & 411.59 & 4.09 & 0.69 & 3.12  \\ 
        Russian & 25,306 & 5,953 & 13,831  & 392.82 & 5.66 & 0.76 & 4.44 & 11,475 & 335.93 &  5.43 & 0.79 & 4.33 \\  
        Latvian & 24,774 &  4,036 & 13,133 & 378.03 & 3.23 & 0.53 & 1.69 & 11,641 & 460.15 & 3.19 & 0.55 & 1.71   \\ \hline
    \end{tabular}}
     \caption{Media partners' datasets used for empirical evaluation of keyword extraction algorithms. }
    \label{tab:media-dataset}
\end{table*}

%\hl{Dodaj:  DODAJ TABELO in opisi za vsak jezik, ali je tagset available ali si ga ti skonsturiral, koliko ima keywordov, ter koliko ima keywordov po zdruzevanju v tocki 3.2}

%\section{Preprocessing}
%\textcolor{green}{Cela ta sekcija opisuje samo preprocessing za TFIDF, tako da to daj pod Section 4 TF-IDF tagset matching enrichment method. Jaz sploh nebi razbil na 3.1 in 3.2, naredi samo dva odstavka  BK: bom dodal tole v TFIDF}
%\label{sec:prep}

\section{Methodology}
\label{sec:methodology}

The recent supervised neural methods are very precise, but, as was already mentioned in Section \ref{sec:intro}, in same cases they do not return a sufficient number of keywords. This is due to the fact that the methods are trained on the training data with a low number of gold standard keywords (as it can be seen from Table \ref{tab:media-dataset}). To meet the media partners' needs, we designed a method that complements state-of-the-art neural methods (the TNT-KID method \cite{martinc2020tntkid} and the transformer-based method proposed by \citet{sahrawat2019keyphrase}, which are both described in Section \ref{sec:related-work}) by a tagset matching approach, returning constant number of keywords ($k$=10). 

\subsection{Transformer-based Keyword Extraction}

Both supervised neural approaches employed in this study are based on the Transformer architecture \citep{vaswani2017attention}, which was somewhat adapted for the specific task at hand. Both models are fed lowercased text consisting of the title and the body of the article. Tokenization is conducted by either using the default BERT tokenizer (when BERT is used) or by employing Sentencepiece tokenizer \cite{kudo2018sentencepiece} (when TNT-KID is used). While the multilingual BERT model is already pretrained on a large corpus consisting of Wikipedias of about 100 languages \citep{devlin2018bert}, TNT-KID requires an additional language model pretraining on the domain specific corpus.    

\subsection{TF-IDF(tm) Tagset Matching}

In our approach, we first take the keywords returned by a neural keyword extraction method and next complement the returned keyword list by adding the missing keywords to achieve the set goal of $k$ keywords. The added keywords are selected by taking the top-ranked candidates from the TF-IDF tagset matching extraction conducted on the preprocessed news articles and keywords. 

\subsubsection{Preprocessing}

\begin{figure}[h]
\resizebox{\linewidth}{!}{\includegraphics{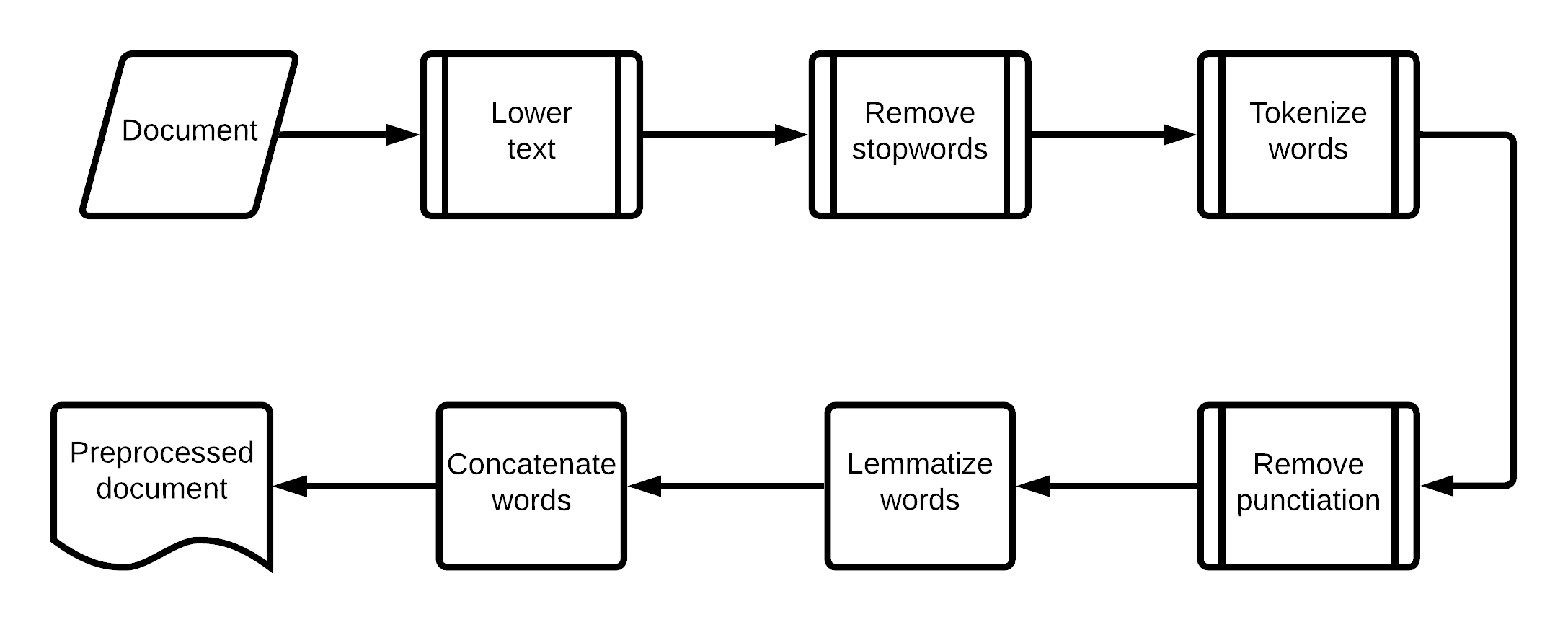}}
\caption{Preprocessing pipeline used for the document normalization and cleaning.} 
\label{fig:prep}
\end{figure}

First, we concatenate the body and the title of the article. After that we lowercase the text and remove stopwords. Finally, the text is tokenized and lemmatized with the Lemmagen3 lemmatizer \citep{jurvsic2010lemmagen}, which supports lemmatization for all the languages except Latvian. For Latvian we use the LatvianStemmer \footnote{\url{https://github.com/rihardsk/LatvianStemmer}}. For the stopword removal we used the \textit{Stopwords-ISO} \footnote{\url{https://github.com/stopwords-iso}} Python library which contained stopwords for all four languages. The final cleaned textual input consists of the concatenation of all of the preprocessed words from the document. We apply the same preprocessing procedure on the predetermined tagsets for each language. The preprocessing procedure is visualized in Figure \ref{fig:prep}.

%\hl{a stopworde torej ven meces iz keywordov - kaj se zgodi z multiword keywords npr. zavod za zaposlovanje - ZA je stopword? Zdaj glavno, da je opisano tako, kot je, je pa za preverit, kako to vpliva;}
%\hl{ BK: Dejansko ne, razen ce je zavod za zaposlovanje tolk pogost da bo stopword, matchal sem full-string-matching ne subsequence. In se en point: Ce je stopowrd in ga ne pobrisem -> tf-idf bo ga prevec penaliziral z idf in bo premajhen in kot tak bo cist zadnje evalviran kot tag  }
%. %BK-FIXED:  \hl{ta figure verjetno sodi bolj pod 3.1}

\subsubsection{TF-IDF Weighting Scheme}
\label{sec:tf-idf}

The TF-IDF weighting scheme \cite{tfidf} assigns each word its weight $w$ based on the frequency of the word in the document (term frequency) and the number of documents the word appears in (inverse document frequency). More specifically, TF-IDF is calculated with the following equation:
   
\begin{equation*}
     \begin{aligned}
         TF-IDF\textsubscript{i} = tf\textsubscript{i,j} \cdot \log_e(\frac{|D|}{df_{i}})
     \end{aligned}
\end{equation*}

The formula has two main components:
\begin{itemize}
 \item  \textit{Term-frequency} (tf) that counts the number of appearances of a word in the document (in the equation above, $tf\textsubscript{i,j}$ denotes  the number of occurrences of the word $i$ in the document $j$)
 \item  \textit{Inverse-document-frequency} (idf) ensures that words appearing in more documents are assigned lower weights (in the formula above $df_{i}$ is the number of documents containing word $i$ and $|D|$ denotes the number of documents).
\end{itemize}

The assumption is that words with a higher TF-IDF value are more likely to be keywords.

\subsection{Tagset Matching Keyword Expansion}

For a given neural keyword extraction method \textit{N}, and for each document \textit{d}, we select $l$ best ranked keywords according to the TF-IDF(tm), which appear in the keyword tagset for each specific dataset. Here, \textit{l} corresponds to \textit{k} - \textit{m}, where $k=10$ and \textit{m} corresponds to the number of keywords returned by a neural method.

Since some of the keywords in the tagsets provided by the media partners were variations of the same root word (i.e., keywords are not lemmatized), we created a mapping from a root word (i.e., a word lemma or a stem) to a list of possible variations in the keyword dataset. For example, a word \textit{'riigieksam'} (\textit{'exam'}) appearing in the article, could be mapped to three tags in the tagset by the Estonian media house with the same root form \textit{'riigieksam'}: \textit{'riigieksamid', 'riigieksamide'} and \textit{'riigieksam'}.

We tested several strategies for mapping the occurrence of a word in the news article to a specific tag in the tagset. For each lemma that mapped to multiple tags, we tested returning a random tag, a tag with minimal length and a tag of maximal length. In the final version, we opted to return the tag with the minimal length, since this tag corresponded to the lemma of the word most often.

\section{Experimental Settings}
\label{sec:media-experiments}

We conducted experiments on the datasets described in Section \ref{sec:dataset}. We evaluate the following methods and combinations of methods: 

\begin{itemize}
\item \textbf{TF-IDF(tm):} Here, we employ the preprocessing and TF-IDF-based weighting of keywords described in Section \ref{sec:methodology} and select the top-ranked keywords that are present in the tagset. 

\item \textbf{TNT-KID} \cite{martinc2020tntkid}: For each dataset, we first pretrain the model with an autoregressive language model objective. After that, the model is fine-tuned on the same train set for the keyword extraction task. Sequence length was set to 256, embedding size to 512 and batch size to 8, and we employ the same preprocessing as in the original study \citep{martinc2020tntkid}.

\item \textbf{BERT + BiLSTM-CRF} \citep{sahrawat2019keyphrase}: We employ an uncased multilingual BERT\footnote{More specifically, we use the 'bert-base-multilingual-uncased' implementation of BERT from the Transformers library (\url{https://github.com/huggingface/transformers}).} model with an embedding size of 768 and 12 attention heads, with an additional BiLSTM-CRF token classification head, same as in \citet{sahrawat2019keyphrase}.

\item \textbf{TNT-KID \& BERT + BiLSTM-CRF}: We extracted keywords with both of the methods and complemented the TNT-KID extracted keywords with the BERT + BiLSTM-CRF extracted keywords in order to retrieve more keywords. Duplicates (i.e., keywords extracted by both methods) are removed. 

\item \textbf{TNT-KID \& TF-IDF}: If the keyword set extracted by TNT-KID contains less than 10 keywords, it is expanded with keywords retrieved with the proposed TF-IDF(tm) approach, i.e., best ranked keywords according to TF-IDF, which do not appear in the keyword set extracted by TNT-KID.

\item \textbf{BERT + BiLSTM-CRF  \& TF-IDF}: If the keyword set extracted by BERT + BiLSTM-CRF contains less than 10 keywords, it is expanded with keywords retrieved with the proposed TF-IDF(tm) approach, i.e., best ranked keywords according to TF-IDF, which do not appear in the keyword set extracted by BERT + BiLSTM-CRF.

\item \textbf{TNT-KID \& BERT + BiLSTM-CRF \& TF-IDF:} the keyword set extracted with the TNT-KID is complemented by keywords extracted with BERT + BiLSTM-CRF (duplicates are removed). If after the expansion the keyword set still contains less than 10 keywords, it is expanded again, this time with keywords retrieved by the TF-IDF(tm) approach.

\end{itemize} 

For TNT-KID, which is the only model that requires language model pretraining, language models were trained on train sets in Table \ref{tab:media-dataset} for up to ten epochs. Next, TNT-KID and BERT + BiLSTM-CRF  were fine-tuned on the training datasets, which were randomly split into 80 percent of documents used for training and 20 percent of documents used for validation. The documents containing more than 256 tokens are truncated, while the documents containing less than 256 tokens are padded with a special $<\textrm{pad}>$ token at the end. We fine-tuned each model for a maximum of 10 epochs and after each epoch the trained model was tested on the documents chosen for validation. The model that showed the best performance on this set of validation documents (in terms of F@10 score) was used for keyword detection on the test set.

\begin{table*}[t]
    \centering
    \resizebox{\linewidth}{!}{
    \begin{tabular}{| c | c | c | c | c | c | c |}
        \hline
        \textbf{Model} & \textbf{P@5} & \textbf{R@5} & \textbf{F1@5} & \textbf{P@10} & \textbf{R@10} & \textbf{F1@10} \\ \hline
                \multicolumn{7}{|c|} {\textbf{Croatian}}\\\hline
        TF-IDF(tm) & 0.2226 &  0.4543 & 0.2988 & 0.1466 & 0.5888 &  0.2347 \\
        TNT-KID & 0.3296 & 0.5135 & 0.4015 & 0.3167 & 0.5359 & 0.3981    \\
        BERT + BiLSTM-CRF &  \textbf{0.4607} & 0.4672 & \textbf{0.4640} &  \textbf{0.4599} &  0.4708 & \textbf{0.4654} \\ 

        TNT-KID \& TF-IDF(tm) & 0.2659 & 0.5670 & 0.3621 & 0.1688 & 0.6944 & 0.2716 \\ 
        BERT + BiLSTM-CRF \& TF-IDF(tm) & 0.2644 & 0.5656 & 0.3604 & 0.1549 & 0.6410 & 0.2495 \\ 
        TNT-KID \& BERT + BiLSTM-CRF &  0.2940  & 0.5447 &  0.3820 & 0.2659 & 0.5968 & 0.3679 \\

        TNT-KID \& BERT + BiLSTM-CRF \& TF-IDF(tm) &  0.2648 & \textbf{0.5681} & 0.3612 & 0.1699 & \textbf{0.7040} & 0.2738\\ \hline

        %\multicolumn{7}{|c|} {\textbf{Croatian}}\\\hline
        %TF-IDF & 0.3430 &  0.4955 & 0.4054 & 0.3364 & 0.4987 &  0.4018\\
        %TNT-KID & 0.3364 & 0.4925 & 0.3998 & 0.3273 & 0.5089 & 0.3984    \\
        %BERT + BiLSTM-CRF &  \textbf{0.4737} & 0.4580 & \textbf{0.4657} &  \textbf{0.4733} &  0.4607 & \textbf{0.4669} \\ 

        %TNT-KID \& TF-IDF(tm) & 0.2835 & 0.5664 & 0.3779 & 0.2594 & 0.6224 & 0.3662\\ 
        %BERT + BiLSTM-CRF \& TF-IDF(tm) & 0.3003 & 0.5569 & 0.3901 & 0.2782 & 0.5732 & 0.3746 \\ 
        %TNT-KID \& BERT + BiLSTM-CRF &  0.2961 & 0.5354 & 0.3813 & 0.2778 & 0.5778 & 0.3752 \\ 

        %TNT-KID \& BERT + BiLSTM-CRF \& TF-IDF(tm) &  0.2653 & \textbf{0.5685} &  0.3618 & 0.2295 & \textbf{0.6549} & 0.3399\\ \hline

           \multicolumn{7}{|c|} {\textbf{Estonian}}\\\hline
        TF-IDF(tm) & 0.0716 &  0.1488 & 0.0966 & 0.0496 & 0.1950 & 0.0790 \\
        TNT-KID & \textbf{0.5194} & 0.5676 & \textbf{0.5424} & \textbf{0.5098} & 0.5942 & \textbf{0.5942}    \\
        BERT + BiLSTM-CRF &  0.5118 & 0.4617 & 0.4855 & 0.5078 & 0.4775 & 0.4922 \\ 

        TNT-KID \& TF-IDF(tm) & 0.3463 & 0.5997 & 0.4391 &  0.1978 & 0.6541 & 0.3037\\ 
        BERT + BiLSTM-CRF \& TF-IDF(tm) & 0.3175 & 0.4978 &  0.3877 & 0.1789 & 0.5381 & 0.2686\\ 
        TNT-KID \& BERT + BiLSTM-CRF &  0.4421 & 0.6014 & 0.5096 & 0.4028 & 0.6438 & 0.4956 \\ 

        TNT-KID \& BERT + BiLSTM-CRF \& TF-IDF(tm) &  0.3588 & \textbf{0.6206} & 0.4547 & 0.2107 & \textbf{0.6912} & 0.3230 \\ \hline

        \multicolumn{7}{|c|} {\textbf{Russian}}\\\hline
        TF-IDF(tm) & 0.1764 &  0.2314 & 0.2002 & 0.1663 & 0.3350 &  0.2223\\
        TNT-KID & \textbf{0.7108} & 0.6007 & \textbf{0.6512} & \textbf{0.7038} & 0.6250 & \textbf{0.6621}    \\
        BERT + BiLSTM-CRF &  0.6901 & 0.5467 & 0.5467 & 0.6849 & 0.5643 & 0.6187 \\ 

        TNT-KID \& TF-IDF(tm) & 0.4519 & 0.6293 & 0.5261 & 0.2981 & 0.6946 & 0.4172\\ 
        BERT + BiLSTM-CRF \& TF-IDF(tm) & 0.4157 & 0.5728 & 0.4818 & 0.2753 & 0.6378 & 0.3846 \\ 
        TNT-KID \& BERT + BiLSTM-CRF &  0.6226 & 0.6375 & 0.6300 & 0.5877 & 0.6707 & 0.6265 \\ 

        TNT-KID \& BERT + BiLSTM-CRF \& TF-IDF(tm) &  0.4622 & \textbf{0.6527} &  0.5412 & 0.2965 & \textbf{0.7213} & 0.4203\\ \hline

        \multicolumn{7}{|c|} {\textbf{Latvian}}\\\hline
        TF-IDF(tm) & 0.2258 & 0.5035 & 0.3118 & 0.1708 & 0.5965 & 0.2655 \\
        TNT-KID &  0.6089 & 0.6887 & \textbf{0.6464} & 0.6054 & 0.6960 & \textbf{0.6476} \\
        BERT + BiLSTM-CRF & \textbf{0.6215} & 0.6214 & 0.6214 & \textbf{0.6204} & 0.6243 & 0.6223   \\ 

        TNT-KID \& TF-IDF(tm) & 0.3402 & \textbf{0.7934} & 0.4762 & 0.2253 & 0.8653 & 0.3575\\ 
        BERT + BiLSTM-CRF \& TF-IDF(tm) & 0.2985 & 0.6957 & 0.4178 & 0.1889 & 0.7427 & 0.3012 \\ 
        TNT-KID \& BERT + BiLSTM-CRF &  0.4545 & 0.7189 & 0.5569 & 0.4341 & 0.7297 & 0.5443 \\ 

        TNT-KID \& BERT + BiLSTM-CRF \& TF-IDF(tm) &  0.3318 & 0.7852 & 0.4666 & 0.2124 & \textbf{0.8672} & 0.3414\\ \hline

    \end{tabular}}
    \caption{Results on the EMBEDDIA media partner datasets.}
    %\hl{At the end rename NEW SPLIT to Croatian and bold in each column}}
    \label{tab:kw_media_eval}
\end{table*}

\section{Evaluation}
    \label{sec:eval}

For evaluation, we employ precision, recall and F1 score. While F1@10 and recall@10 are the most relevant metrics for the media partners, we also report precision@10, precision@5, recall@5 and F1@5. Only keywords which appear in a text (present keywords) were used as a gold standard, since we only evaluate approaches for keyword tagging that are not capable of finding keywords which do not appear in the text. Lowercasing and lemmatization (stemming in the case of Latvian) are performed on both the gold standard and the extracted keywords (keyphrases) during the evaluation. The results of the evaluation on all four languages are listed in Table \ref{tab:kw_media_eval}.

Results suggest, that neural approaches, TNT-KID and BERT+BiLSTM-CRF offer comparable performance on all datasets but nevertheless achieve different results for different languages. TNT-KID outperforms BERT-BiLSTM-CRF model according to all the evaluation metrics on the Estonian and Russian news dataset. It also outperforms all other methods in terms of precision and F1 score. On the other hand, BERT+BiLSTM-CRF performs better on the Croatian dataset in terms of precision and F1-score. On Latvian TNT-KID achieves top results in terms of F1, while BERT+BiLSTM-CRF offers better precision. 

Even though the TF-IDF tagset matching method performs poorly on its own, we can nevertheless drastically improve the recall@5 and the recall@10 of both neural systems, if we expand the keyword tag sets returned by the neural methods with the TF-IDF ranked keywords. The improvement is substantial and consistent for all datasets, but it nevertheless comes at the expanse of the lower precision and F1 score. This is not surprising, since the final expanded keyword set always returns 10 keywords, i.e., much more than the average number of present gold standard keywords in the media partner datasets (see Table \ref{tab:media-dataset}), which badly affects the precision of the approach. Nevertheless, since for a journalist a manual inspection of 10 keyword candidates per article and manual selection of good candidates (e.g., by clicking on them) still requires less time than the manual selection of keywords from an article, we argue that the improvement of recall at the expanse of the precision is a good trade off, if the system is intended to be used as a recommendation system in the media house environment.   

Combining keywords returned by TNT-KID and  BERT + BiLSTM-CRF also consistently improves recall, but again at the expanse of lower precision and F1 score. Overall, for all four languages, the best performing method in terms of recall is the TNT-KID \& BERT + BiLSTM-CRF \& TF-IDF(tm).

\section{Conclusion and Future Work}
\label{sec:conc}

In this work, we tested two state-of-the-art neural approaches for keyword extraction, TNT-KID \cite{martinc2020tnt} and BERT BiLSTM-CRF \cite{sahrawat2019keyphrase}, on three less resourced European languages, Estonian, Latvian, Croatian, as well as on Russian. We also proposed a tagset based keyword expansion approach, which drastically improves the recall of the method, making it more suitable for the application in the media house environment. 

Our study is one of the very few studies where supervised keyword extraction models were employed on several less resourced languages. The results suggest that these models perform well on languages other than English and could also be successfully leveraged for keyword extraction on morphologically rich languages. 

The focus of the study was whether we can improve the recall of the supervised models, in order to make them more useful as recommendation systems in the media house environment. Our method manages to increase the number of retrieved keywords, which drastically improves the recall for all languages. For example, by combing all neural methods and the TF-IDF based approach, we improve on the recall@10 achieved by the best performing neural model, TNT-KID, by 16.81 percentage points for Croatian, 9.70 percentage points for Estonian, 9.63 percentage points for Russian and 17.12 percentage points for Latvian. The resulting method nevertheless offers lower precision, which we will try to improve in the future work.

In the future we also plan to perform a qualitative evaluation of our methods by journalists from the media houses. Next, we plan to explore how adding background knowledge from knowledge databases - lexical (e.g. Wordnet\cite{wordnet}) or factual (e.g. WikiData\cite{wikidata}) would benefit the aforementioned methods. The assumption is that with the linkage of the text representation and the background knowledge we would achieve a more representative understanding of the articles and the concepts appearing in them, which would result in a more successful keyword extraction. 

In traditional machine-learning setting a common practice of combining different classifier outputs to a single output is referred  to as stacking. We propose further research on this topic by testing combinations of various keyword extraction models. Finally, we also plan to further improve our unsupervised TF-IDF based keyword extraction method. One way to to do this would be to add the notion of positional encoding, since some of the keywords in the news-media domain often can be found at the beginning of the article and the TF-IDF(tm) does not take this into account while applying the weighting on the matched terms.    

%One can explore the possibilities of adding a final supervised layer  
% \label{sec:further_work}
% We want to explore how adding background knowledge from knowledge bases - lexical (e.g. Wordnet\cite{wordnet}) or factual (e.g. WikiData\cite{wikidata}) would benefit the aforementioned methods. The assumption is that with the linkage of the text representation and the background knowledge we would achieve a more representative understanding of the articles and the concepts appearing in them, hoping for a more sucessful keyword extraction.
% \par One way to exploit the lexical based knowledge graphs would be to use them for a synonym based keyword extraction task. For example the WordNet knowledge graph provides a system of synonyms, which when paired with detected keywords from the tagsets would provide a richer user experience for the task of keywords extraction on articles.
% \par Finally, some graph-neural attention based methods can be explored on the linked representations of texts and knowledge bases to try to extract more keywords with prior graph theoretic notion.

\section{Acknowledgements}
This paper is supported by European Union’s Horizon 2020 research and innovation programme under grant agreement No. 825153, project EMBEDDIA (Cross-Lingual Embeddings for Less-Represented Languages in European News Media). The third author was financed via young research ARRS grant.
Finally, the authors acknowledge the financial support from the Slovenian Research Agency for research core funding for the programme Knowledge Technologies (No. P2-0103) and the project TermFrame - Terminology and Knowledge Frames across Languages (No. J6-9372).
\bibliographystyle{acl_natbib}
\bibliography{eacl2021}

\end{document}